\PassOptionsToPackage{greek,english}{babel}
\documentclass[runningheads,a4paper]{llncs}

\usepackage[american]{babel}
\usepackage{graphicx}
\usepackage{array}
\usepackage{setspace}
\usepackage{multirow,mathptmx}
\usepackage{url}
\usepackage[table]{xcolor}
\usepackage{amsmath,amssymb,amsfonts}
\usepackage[utf8]{inputenc}
\usepackage[T1]{fontenc}
\usepackage{algorithm2e}
\usepackage{verbatim}
\usepackage{booktabs}
\usepackage{paralist}
\usepackage{tabularx}

\urldef{\mails}\path|{rbajpai,sporia,danho,cambria}@ntu.edu.sg|

\usepackage{csquotes}

\usepackage[T1]{fontenc}

\usepackage{microtype}

\usepackage[%
rm={oldstyle=false,proportional=true},%
sf={oldstyle=false,proportional=true},%
tt={oldstyle=false,proportional=true,variable=true},%
qt=false%
]{cfr-lm}
\usepackage[math]{blindtext}

\usepackage[
bookmarks=false,
breaklinks=true,
colorlinks=true,
linkcolor=black,
citecolor=black,
urlcolor=black,
pdfpagelayout=SinglePage
]{hyperref}
\usepackage[all]{hypcap}

\usepackage[capitalise,nameinlink]{cleveref}
\crefname{section}{Sect.}{Sect.}
\Crefname{section}{Section}{Sections}
\crefname{figure}{Fig.}{Fig.}
\Crefname{figure}{Figure}{Figures}

\usepackage{xspace}

\newcommand{\makefigure}[3]{
	\begin{figure}[h]
		\begin{center}
			\includegraphics[width=#1mm]{#2}
		\end{center}
		\caption{#3}
		\label{fig:#2}
	\end{figure}
}

\newcolumntype{x}[1]{>{\centering\hspace{0pt}}p{#1}}

\newcommand{\caplab}[2]{\caption{#1}\label{#2}}

\newcommand{\conftbl}{\centering\footnotesize\vspace{5pt}}

\newcommand{\furl}[1]{\footnote{\url{http://#1}}}

\newcommand{\tn}{\tabularnewline\hline}
\DeclareFontFamily{U}{MnSymbolC}{}
\DeclareSymbolFont{MnSyC}{U}{MnSymbolC}{m}{n}
\DeclareFontShape{U}{MnSymbolC}{m}{n}{
    <-6>  MnSymbolC5
   <6-7>  MnSymbolC6
   <7-8>  MnSymbolC7
   <8-9>  MnSymbolC8
   <9-10> MnSymbolC9
  <10-12> MnSymbolC10
  <12->   MnSymbolC12%
}{}
\DeclareMathSymbol{\powerset}{\mathord}{MnSyC}{180}

\makeatletter
\g@addto@macro{\UrlBreaks}{\UrlOrds}
\makeatother

\begin{document}

\input glyphtounicode.tex
\pdfgentounicode=1

\title{Developing a concept-level knowledge base\\for sentiment analysis in Singlish}
\titlerunning{Singlish Sentiment Lexicon}

\author{Rajiv Bajpai \and Soujanya Poria \and Danyun Ho \and Erik Cambria}
\authorrunning{Rajiv Bajpai et al.}
\institute{School of Computer Science and Engineering\\Nanyang Technology University\\\mails}

%

\maketitle

\begin{abstract}
In this paper, we present Singlish sentiment lexicon, a concept-level knowledge base for sentiment analysis that associates multiword expressions to a set of emotion labels and a polarity value.
Unlike many other sentiment analysis resources, this lexicon is not built by manually labeling pieces of knowledge coming from general NLP resources such as WordNet or DBPedia. Instead, it is automatically constructed by applying graph-mining and multi-dimensional scaling techniques on the affective common-sense knowledge collected from three different sources. This knowledge is represented redundantly at three levels: semantic network, matrix, and vector space. Subsequently, the concepts are labeled by emotions and polarity through the ensemble application of spreading activation, neural networks and an emotion categorization model.
\end{abstract}


\section{Introduction}\label{sec:intro}
Singapore Colloquial English (or Singlish), as the name suggests, refers to an English-based creole language, or rather the variety of English spoken in Singapore \cite{brown1999singapore}. A product of colonial implantation, English as spoken by the British is gradually ‘nativized’ into Singlish due to extensive language borrowing and mixing with other language in the linguistic environment of Singapore. The end product of language contact is Singlish, an English variety that shows ‘a high degree of influence from other local languages such as Hokkien, Cantonese, Mandarin, Malay and Tamil \cite{platt1980english}.

\cite{gupta1994step} noted that English in Singapore is diglossic in nature, which its domain of use is dependent on the degree of formality. The high variety is used in formal situations while the low variety used for informal occasions. The high variety of English is called Singapore Standard English and it is like other Standard English varieties (i.e., American English or British English). The low variety is called Singapore Colloquial English, or Singlish. It is distinct from Standard English because it demonstrates unique syntactical and lexical features \cite{low2005english}. \cite{gupta1994step}, however, notes that there is no stark demarcation between Singlish and Singapore Standard English, but rather degree of variations in between.

Singapore is a multi-ethnic/multilingual country consisting of 77\% Chinese, 14\% Malay and 8\% Indian. While Singlish serves as an inter-ethnic lingua franca \cite{harada2009roles}, linguistic boundaries are present in the variety despite extensive contacts between different speakers. \cite{tay1982uses} observed that the speaker’s linguistic background plays a part in influencing the linguistic repertoire of the user when using Singlish. For example a Chinese speaker may exclaim “John sibei hum sup one” whereas a Malay speaker may cry “John very buaya sia”. While both phrases essentially mean “John is so lecherous” in Singlish, one’s choice of lexicon in Singlish might be influenced by one’s language background. The presence of ethnic differentiation therefore shows that Singlish, is not a ‘monolithic’ linguistic entity spoken by a homogenous Singaporean ‘ethnicity’. 

People in Singapore tend to speak and write in Singapore Colloquial English. Due to the informal nature of Singlish, it has found its way on the Internet as Singaporeans prefer to use Singlish in informal communication like online chatting, tweeting or interaction on social media like Facebook \cite{warschauer2002internet}. As such, large amount of data in Singlish is inaccessible because it is a variety that is highly distinct and incomprehensible to English speakers elsewhere. Researchers have been studying the linguistic features of Singlish extensively since the 1960s, but little research on Singlish has been done in natural language processing (NLP). Sentiment analysis \cite{pang2008opinion,zadeh2016multimodal} and using multimodal information as a new trend  \cite{zadeh2016mosi,poria2015deep,traffickingacl17,soujanyaacl17,poria2016convolutional} is a popular branch of NLP research that aims to understand sentiment of documents automatically using combination of various machine learning approaches \cite{zadeh2015micro,tensoremnlp17,soujanyaacl17,ceclm17}. 
A sub-branch of NLP research, \cite{phua2013social} conducted a study on sentiment analysis in Singapore Colloquial English. Using 425 Singlish posts on a political issue in Singapore, the baseline of sentiment analysis was determined by manually annotating the target phrases with either a positive or negative polarity. The researcher then ran a phrasal sentiment analysis classifier to analyze the polarity of the dataset. An overall accuracy of 35.5\% was achieved by this approach, in which a precision of 21\% was achieved for positive posts and a precision of 94.4\% was achieved for negative post.

\section{AffectNet and ConceptNet}
As an inventory of target labels and a source of training examples for the supervised classification, we used the emotion lists  provided for the SemEval 2007 task 14: Affective text. According to the organizers of this task, the lists were extracted from WNA \cite{strapparava2004wordnet}. There are six lists corresponding to the six basic emotions: anger, fear, disgust, sadness, surprise, and joy. This dataset assigns emotion labels to synsets groups of words or concepts that are synonymous in the corresponding senses: e.g., a synset {puppy love, calf love, crush, infatuation} is assigned the label JOY. However, we ignored the synonymy information contained in the data and used the labels for individual words or concepts, i.e., puppy love$\rightarrow$JOY, calf love$\rightarrow$JOY, crush$\rightarrow$JOY, infatuation$\rightarrow$JOY.
ConceptNet represents the information from the Open Mind corpus as a directed
graph, in which the nodes are concepts and the labeled edges are common-sense assertions that interconnect them \cite{camcom}.

\section{AffectiveSpace}
\label{affspace}
AffectiveSpace \cite{cambria2015AffectiveSpace} is a novel affective common-sense knowledge visualization and analysis system. 
The human mind constructs intelligible meanings by continuously compressing over vital relations \cite{fauconnier2008way}.The compression principles aim to transform diffuse and distended conceptual structures to more focused versions so as to become more congenial for human understanding. 
To this end, principal component analysis (PCA) has been applied on the matrix representation of AffectNet. 
In particular, truncated singular value decomposition (TSVD) has been preferred to other dimensionality reduction techniques for its simplicity, relatively low computational cost, and compactness. 
TSVD, in fact, is particularly suitable for measuring the cross-correlations between affective common-sense concepts as it uses an orthogonal transformation to convert the set of possibly correlated common-sense features associated with each concept into a set of values of uncorrelated variables (the principal components of the SVD). 

By using Lanczos' method \cite{lanczos1950iteration}, moreover, the
generalization process is relatively fast (a few seconds), despite the size and the sparseness of AffectNet.
The objective of such compression is to allow many details in the blend of ConceptNet and WNA to be removed such that the blend only consists of a few essential features that represent the global picture. Applying TSVD on AffectNet, in fact, causes it to describe other features that could apply to known affective concepts by analogy: if a concept in the matrix has no value specified for a feature owned by many similar concepts, then by analogy the concept is likely to have that feature as well. In other words, concepts and features that point in similar directions and, therefore, have high dot products, are good candidates for analogies.

A pioneering work on understanding and visualizing the affective information associated with natural language text was conducted by Osgood et al.~\cite{osgood1975cross}. Osgood used multi-dimensional scaling (MDS) to create visualizations of affective words based on similarity ratings of the words provided to subjects from different cultures. 
Words can be thought of as points in a multi-dimensional space and the similarity ratings represent the distances between these words. MDS projects these distances to points in a smaller dimensional space (usually two or three dimensions).
Similarly, AffectiveSpace aims to grasp the semantic and affective similarity between different concepts by plotting them into a multi-dimensional vector space \cite{cambria2010sentic}. 
Unlike Osgood's space, however, the building blocks of AffectiveSpace are not simply a limited set of similarity ratings between affect words, but rather millions of confidence scores related to pieces of common-sense knowledge linked to a hierarchy of affective domain labels \cite{poremo}. 
Rather than merely determined by a few human annotators and represented as a word-word matrix, in fact, AffectiveSpace is built upon an affective common-sense knowledge base, namely AffectNet, represented as a concept-feature matrix.
After performing TSVD on such matrix, hereby termed $A$ for the sake of conciseness, a low-rank approximation of it is obtained, that is, a new matrix $\tilde{A} = U_k \, \Sigma_k \, V^T_k$.

This approximation is based on minimizing the Frobenius norm of the difference between $A$ and $\tilde{A}$ under the constraint $rank(\tilde{A}=k$.
For the Eckart--Young theorem \cite{eckart1936approximation},it represents the best approximation of $A$ in the least-square sense:

\begin{equation}
\min_{\tilde{A}|rank(\tilde{A}=k} | A - \tilde{A} | = \min_{\tilde{A}|rank(\tilde{A}=k} | \Sigma - U^*\tilde{A}V | = \min_{\tilde{A}|rank(\tilde{A}=k} | \Sigma - S | 
\end{equation}

assuming that $\tilde{A}$ has the form $\tilde{A} = USV^*$, where $S$ is diagonal. 
From the rank constraint, i.e., $S$ has $k$ non-zero diagonal entries, the minimum of the above statement is obtained as follows:
\begin{equation}
\min_{\tilde{A}|rank(\tilde{A}=k} | \Sigma - S | = \min_{s_i}\sqrt{\sum_{i=1}^{n}{(\sigma_i-s_i)^2}}
\end{equation}
\begin{equation}
\min_{s_i}\sqrt{\sum_{i=1}^{n}{(\sigma_i-s_i)^2}} = \min_{s_i}\sqrt{\sum_{i=1}^{k}{(\sigma_i-s_i)^2}+\sum_{i=k+1}^{n}{\sigma_i^2}} = \sqrt{\sum_{i=k+1}^{n}{\sigma_i^2}}
\end{equation}

Therefore, $\tilde{A}$ of rank $k$ is the best approximation of $A$ in the Frobenius norm sense when $\sigma_i=s_i$ $(i=1,...,k)$ and the corresponding singular vectors are the same as those of $A$.
If all but the first $k$ principal components are discarded, common-sense concepts and emotions are represented by vectors of $k$ coordinates.
These coordinates can be seen as describing concepts in terms of `eigenmoods'
that form the axes of AffectiveSpace, i.e., the basis $e_0$,...,$e_{k-1}$ of the vector space. 
For example, the most significant eigenmood, $e_0$, represents concepts with positive affective valence.
That is, the larger a concept's component in the $e_0$ direction is, the more affectively positive it is likely to be.

Concepts with negative $e_0$ components, then, are likely to have negative affective valence.
Thus, by exploiting the information sharing property of TSVD, concepts with the same affective valence are likely to have similar features -- that is, concepts conveying the same emotion tend to fall near each other in AffectiveSpace.
Concept similarity does not depend on their absolute positions in the vector space, but rather on the angle they make with the origin.
For example, concepts such as \texttt{gei\_yan}, \texttt{see\_buay}, and \texttt{cham\_sheung} are found very close in direction in the vector space, while concepts like \texttt{act\_blur}, \texttt{ah\_long}, and \texttt{bo\_chap} are found in a completely different direction (nearly opposite with respect to the centre of the space).

\section{The Emotion Categorization Model}
\label{ecm}
The Hourglass of Emotions \cite{camsenold} is an affective categorization model inspired by Plutchik's studies on human emotions \cite{plunat}.
It reinterprets Plutchik's model by organizing primary emotions around four independent but concomitant dimensions, whose different levels of activation make up the total emotional state of the mind. Such a reinterpretation is inspired by Minsky's theory of the mind, according to which brain activity consists of different independent resources and that emotional states result from turning some set of these resources on and turning another set of them off. 

This way, the model can potentially synthesize the full range of emotional experiences in terms of Pleasantness, Attention, Sensitivity, and Aptitude, as the different combined values of the four affective dimensions can also model affective states we do not have a specific name for, due to the ambiguity of natural language and the elusive nature of emotions. 
The primary quantity we can measure about an emotion we feel is its strength. But, when
we feel a strong emotion, it is because we feel a very specific emotion. And, conversely, we cannot feel a specific emotion like fear or amazement without that emotion being reasonably strong. 
For such reasons, the transition between different emotional states is modeled, within the same affective dimension, using the function $G(x) = -\frac{1}{{\sigma \sqrt {2\pi } }}e^{{{ - x^2 } \mathord{\left/ {\vphantom {{ - \left( {x - \mu } \right)^2 } {2\sigma ^2 }}} \right. \kern-\nulldelimiterspace} {2\sigma ^2 }}}$, for its symmetric inverted bell curve shape that quickly rises up towards the unit value. 
In particular, the function models how valence or intensity of an affective dimension varies according to different values of arousal or activation ($x$), spanning from null value (emotional void) to the unit value (heightened emotionality).
Justification for assuming that the Gaussian function (rather than a step or simple linear function) is appropriate for modeling the variation of emotion intensity is based on research into the neural and behavioral correlates of emotion, which are assumed to indicate emotional intensity in some sense. 
Nobody genuinely knows what function subjective emotion intensity follows, because it has never been truly or directly measured. 



Each affective dimension of the Hourglass model is characterized by six levels of activation (measuring the strength of an emotion), termed `sentic levels', which represent the intensity thresholds of the expressed or perceived emotion \cite{camact}.
These levels are also labeled as a set of 24 basic emotions, six for each of the affective dimensions, in a way that allows the model to specify the affective information associated with text both in a dimensional and in a discrete form.
The dimensional form, in particular, is termed `sentic vector' and it is a four-dimensional $float$ vector that can potentially synthesize the full range of emotional experiences in terms of Pleasantness, Attention, Sensitivity, and Aptitude. 
In the model, the vertical dimension represents the intensity of the different affective dimensions, i.e., their level of activation, while the radial dimension represents K-line that can activate configurations of the mind, which can either last just a few seconds or years.
The model follows the pattern used in color theory and research in order to obtain judgments about combinations, i.e., the emotions that result when two or more fundamental emotions are combined, in the same way that red and blue make purple.
Hence, some particular sets of sentic vectors have special names, as they specify well-known compound emotions. 
For example, the set of sentic vectors with a level of Pleasantness~$\in$~[G(2/3), G(1/3)), i.e., joy, a level of Aptitude $\in$~[G(2/3), G(1/3)), i.e., trust, and a minor magnitude of Attention and Sensitivity, are termed `love sentic vectors' since they specify the compound emotion of love. 
More complex emotions can be synthesized by using three, or even four, sentic levels, e.g., joy + trust + anger = jealousy.
Therefore, analogous to the way primary colors combine to generate different color gradations (and even colors we do not have a name for), the primary emotions of the Hourglass model can blend to form the full spectrum of human emotional experience.

\begin{table}[h]\conftbl
	\caplab{Sensitivity cluster}{tbl:s-cluster}
	\begin{tabular}{|p{2.5cm}|p{2.5cm}|p{6.7cm}|}\hline
		{\bf Rage} &{Vomit Blood}&{Used to expressed something that is extremely aggravating, frustrating or difficult to endure, word-for-word translation. Loan translation from Chinese}\tn
		{\bf Anger}&{Tu Lan}&{Used to express something that is extremely aggravating, frustrating or difficult to endure. Probably borrowed from Hokkien}\tn
		{\bf Frustration}&{Pek Cek}&{To denote exasperation or frustration. Probably borrowed from Hokkien
		}\tn
		{\bf Terror}&{Balls Drop}&{To denote being frightened, scared or shocked. }\tn
		{\bf Fear}&{Khia-khia}&{To denote being afraid, worried or nervous. Literally “scared scared”. Probably borrowed from Hokkien}\tn
		{\bf Apprehension}&{Kan Cheong}&{To denote being nervous or tense. Probably borrowed from Cantonese}\tn
	\end{tabular}
\end{table}
\begin{table}[h]\conftbl
	\caplab{Attention cluster}{tbl:a-cluster}
	\begin{tabular}{|p{2.5cm}|p{2.5cm}|p{6.8cm}|}\hline
		{\bf Vigilance} &{On the ball }&{To describe someone who is alert, hardworking and enthusiastic}\tn
		{\bf Anticipation}&{Can't Wait}&{Same as “Can’t wait” or “looking forward to” in Standard English}\tn
		{\bf Interest}&{Enthu}&{Showing or having eager enjoyment or interest. Truncation of the word ‘Enthusiastic’ 
		}\tn
		{\bf Amazement}&{Wah}&{Interjection to denote admiration or awe. Probably borrowed from Chinese}\tn
		{\bf Surprise}&{Alamak}&{Interjection to denote surprise and shock. Probably borrowed from Malay.}\tn
		{\bf Distinction}&{Itchy Finger}&{To denote restlessness or a restive person who does something disruptive out of boredom.
		}\tn
	\end{tabular}
\end{table}
\begin{table}[h]\conftbl
	\caplab{Pleasantness cluster}{tbl:p-cluster}
	\begin{tabular}{|p{2.5cm}|p{2.5cm}|p{6.8cm}|}\hline
		{\bf Ecstasy} &{Shiok}&{To denote sheer delight. Probably borrowed from Malay “syok” or from Punjabi “shauk”}\tn
		{\bf Joy}&{Happy like bird}&{To denote one who is overly happy or overjoyed.}\tn
		{\bf Serenity}&{Chill}&{To describe someone who is relaxed or easy-going. From North American informal usage but appears frequently in Singlish context.
		}\tn
		{\bf Grief}&{Sim Tia / Heart pain}&{Denotes heartache. Literally “heart pain”. Probably borrowed from Hokkien}\tn
		{\bf Sadness}&{Sad}&{Same as “sad” in Standard English
		}\tn
		{\bf Pensiveness}&{Emo}&{Short form of ‘emotional’, to denote moodiness.}\tn
	\end{tabular}
\end{table}
\begin{table}[h]\conftbl
	\caplab{Aptitude cluster}{tbl:ap-cluster}
	\begin{tabular}{|p{2.5cm}|p{2.3cm}|p{7cm}|}\hline
		{\bf Admiration} &{Suka}&{To like or be fond of. Probably borrowed from Malay}\tn
		{\bf Trust}&{Trust}&{Same as “trust” in Standard English
		}\tn
		{\bf Acceptance}&{Accept}&{Same as “accept” in Standard English
		}\tn
		{\bf Loathing
		}&{Hate}&{Same as “hate” in Standard English
	}\tn
	{\bf Disgust}&{Erxin}&{To denote disgust. Probably borrowed from Mandarin}\tn
	{\bf Boredom}&{Sian}&{To denote boredom and frustration. Probably borrowed from Hokkien}\tn
\end{tabular}
\end{table}

Beyond emotion detection, the Hourglass model is also used for polarity detection tasks. 
Since polarity is strongly connected to attitudes and feelings, it is defined in terms of the four affective dimensions, according to the formula:
\begin{equation}
p=\sum_{i=1}^{N}\frac{{Pleasantness(c_i)+|Attention(c_i)|-|Sensitivity(c_i)|+Aptitude(c_i)}}{3N}
\label{eq:pol}
\end{equation}

where $c_i$ is an input concept, $N$ the total number of concepts, and 3 the normalization factor (as the Hourglass dimensions are defined as float $\in$ [-1,+1]). 
In the formula, Attention is taken as absolute value since both its positive and negative intensity values correspond to positive polarity values (e.g., `surprise' is negative in the sense of lack of Attention, but positive from a polarity point of view). 

Similarly, Sensitivity is taken as negative absolute value since both its positive and negative intensity values correspond to negative polarity values (e.g., `anger' is positive in the sense of level of activation of Sensitivity, but negative in terms of polarity). Besides practical reasons, the formula is important because it shows a clear connection between polarity (opinion mining) and emotions (sentiment analysis) \cite{camacsa}.


\section{Developing Singlish Sentiment Lexicon}
In this section, we elaborate the development of Singlish Sentiment Lexicon based on the original English SenticNet \cite{camsen,cambria2016senticnet} and its extensions \cite{porenr,pormer}. We first explain the characteristics of Singlish language and concepts before going to the detail of the proposed method.
\subsection{Syntactic Features of Singlish Language}
\label{Syntax Singlish}
This section contains a review of Singlish syntactical features that are relevant to the semantic analysis of the language. While these features characterize the syntax of Singlish, they are by no means compulsory in the grammatical construct of the language. The seemingly free variation of Singlish syntax could be attributed to the syntactical competition between British English, the suprastrate language that forms the grammatical base of Singlish, and substrate languages like Hokkien and Malay that influences the grammatical features of Singlish \cite{satrad}.

\subsubsection{Noun and pronoun omission}
Q: Why does Erik like to eat laksa?

A: Erik likes to eat laksa because it is delicious

\begin{enumerate}
	\item{\O\,likes to eat laksa because it is delicious.}
	\item{Erik likes to eat \O\,because it is delicious.}
	\item{Erik likes to eat laksa because \O\,is delicious.}
\end{enumerate}
\subsubsection{Copula dropping}
\begin{enumerate}
	\item{Eric \O\,very clever (Eric is very clever)}
	\item{Nick \O\,going to church (Nick is going to church)}
	\item{That one \O\,Rachel husband. (That one is Rachel’s husband)}
	\item{My house \O\,in Serangoon Gardens (My house is in Serangoon Gardens)}
	\item{The dog \O\,hit by a car (The dog is hit by a car)}
\end{enumerate}

The copular verb “to be” can be omitted in several instances in Singlish. It is usually omitted when the subject occurs with an adjective phrase containing the adverb “very” or “so” (2a), when it is before the present participle (-ing form) of a verb (2b), after a noun phrase (2c) or a locative word (2d), and in the passive construction (2e).

\subsection{Lexical features of Singlish}
A review on the lexical features of Singlish is conducted. They could be categorized in three distinct forms, namely borrowing from non-English languages, English words that are “nativized” to the Singapore context, and English words that are similar to other English varieties. 

The first source of Singlish originates from the borrowing of expressions from non-English languages spoken in Singapore. Singlish words of non-English origins include “shiok” that conveys sheer delight (from Malay “syok” or Punjabi “shauk”), “kan cheong” that denotes someone who is nervous or tense (from Cantonese), “sian” that refers to a state of boredom (from Hokkien) and more.

The second source of Singlish is English words that have different meaning in Singlish. The implantation of English in a different language setting has led to the idiosyncratic use of English words that are meaningful to the Singlish context. These English-origin Singlish words have meaning and usage that are different to Standard English, and they are created through the sharing of different ideals and expressions between cultures in a novel setting.
These sources of words come from calques (word for word translation) like “vomit blood” from Chinese (literally “to vomit blood), semantic shift (change in meaning) like “steady” which refers to someone who is calm and collected (rather than “firmly balanced” in Standard English) and the coining of idiosyncratic metaphors like “happy like bird” which denotes a state of great joy. 

\makefigure{79}{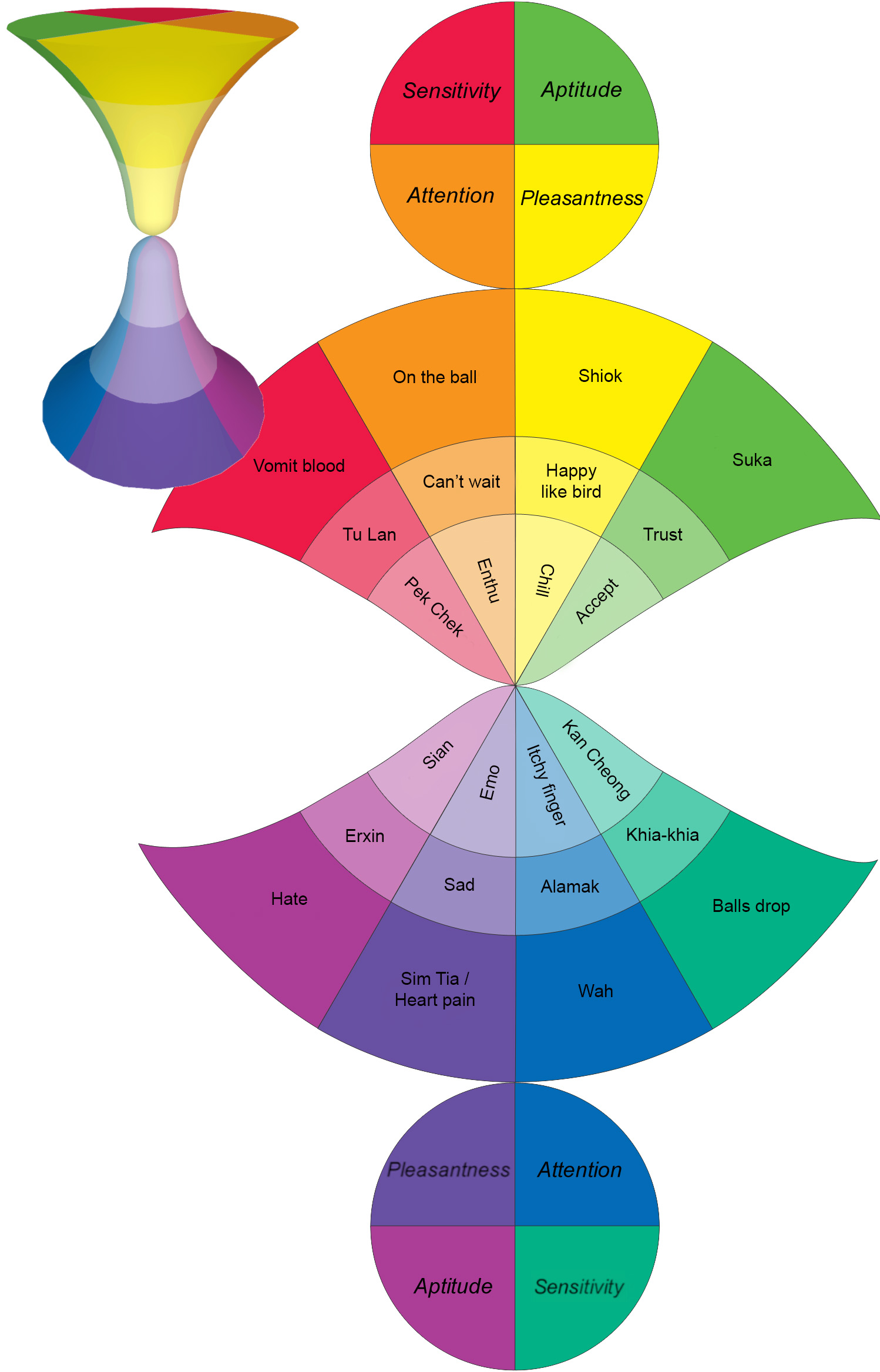}{\small{Singlish Concepts Projected into Hourglass of Emotion}}

The third source of Singlish is from Standard English, and they have same meaning (like ‘hate’) like other varieties of English elsewhere. As much as the lexical composition of Singlish owes to the borrowing of other languages of Singapore, English still serves as the lexifier, or rather the dominant language where most of the vocabulary of Singlish originates. These English words in Singlish are either inherited from English, or borrowed from other English varieties, like “chill” from American informal usage.

\subsection{Hourglass of Emotion Model and Singlish Emotion Expression}
A survey on emotional expressions of Singlish is conducted by reviewing Singlish wordlists from the Coxford Singlish Dictionary  and the Singlish Dictionary \cite{lee2011unsupervised}, and also the knowledge of the one of the researchers who is a native speaker of Singlish. 
Figure 1 shows how the different emotion terms fit into the emotional categories of the model. These emotional expressions commonly appear in Singlish discourse, and they are made up of loanwords from other languages, English words with ‘idiosyncractic’ meaning, and English words. The model, as far as possible, include Singlish emotion words that describe general emotional states, like ‘tu lan’ for anger. Singlish emotion words with specific emotion referents are excluded, like ‘jelat’ that refers to a sense of disgust that pertains to satiated by rich tasting food to the point that one is sick of it. 

Figure 1 is not the translated equivalent of hourglass model as they contain Singlish emotion expressions that best fit into the model. As such, some of the emotion terms are not universally known. Chinese-origin words in Singlish like ‘tu lan’, ‘pek chek’ and ‘erxin’ are mostly used by Chinese Singaporeans and they might not be used by Singaporeans of other ethnicities. Some of the emotion terms might have restricted or emotional domains. For example the word ‘sian’ which usually refers to a state of boredom also has a secondary meaning of frustration.


\subsection{The Method}
In this section, we propose a simple but effective method to construct Singlish sentiment lexicon based on the already developed English SenticNet as benchmark. This process involves following fundamental steps - 

\begin{itemize}
	\item Represent ConceptNet and Singlish concepts into AffectNet structure. 
	\item Blend Singlish AffectNet and AffectNet with ConceptNet to obtain the new AffectiveSpace.
	\item After employing TSVD each Singlish concepts are represented by 100 dimensional vector.
	\item These vectors are used as features for a supervised classifier which is trained using English SenticNet benchmark.
\end{itemize}

Each Singlish concept has been labeled manually into one of the emotion categories (Anger, Disgust, Surprise, Joy, Sadness, Fear). Then we create the Singlish AffectNet graph in the following way - 
\begin{itemize}
	\item Take the emotions, i.e., Anger, Disgust Surprise, Joy, Sadness, Fear as the nodes in the graph.
	\item For each Singlish concept draw an edge from that concept to the emotion node in step 1. Each of such edge is labeled by "HasProperty".
\end{itemize}

ConceptNet is represented in the form of a labeled direct graph, with nodes being concepts such as, for example, spoon, eating, book, paper, and arcs being relations such as UsedFor (spoon--UsedFor$\rightarrow$eating) and MadeOf (book--MadeOf$\rightarrow$paper).
Technically, a graph can be thought of as a matrix. 

\begin{table}
\definecolor{ColumnColor}{rgb}{1,0.9451,0}
\definecolor{RowColor}   {rgb}{1,0.9686,0.6039}
\begin{tabular}{|x{90pt}|x{15pt}>{\columncolor{ColumnColor}}x{43pt}x{43pt}x{55pt}x{66pt}c|}
\hline
Concepts & \multicolumn6{c|}{Semantic Features}                 \\
        & \multicolumn6{c|}{({\it relationship+concept})} \\
\cline{2-7}
& $\dots$ & {\it Causes}\\shiok & {\it IsA}\\event  & {\it UsedFor}\\cooking & {\it MotivatedBy}\\celebration  & $\dots$ \\
\hline
                    $\vdots$        &         & $\vdots$ & $\vdots$ & $\vdots$ & $\vdots$ &         \\
\rowcolor{RowColor} hari\_raya            & $\dots$ &     x    &     x    &     --     &     x    & $\dots$ \\
                       makan         & $\dots$ &     x     &    --   &    --    &      --    & $\dots$ \\
                       bamboo\_steamer         & $\dots$ &     --     &     --   &     x    &      --    & $\dots$ \\
                    \bf special\_occasion   & $\dots$ & \bf x?   & \bf x    & \bf  --    & \bf x    & $\dots$ \\
\rowcolor{RowColor} birthday        & $\dots$ &     x    &     x    &     --     &     x    & $\dots$ \\
                    $\vdots$        &         & $\vdots$ & $\vdots$ & $\vdots$ & $\vdots$ &         \\
\hline
\end{tabular}
\caplab{Cumulative analogy allows for the inference of new pieces of knowledge by comparing similar concepts. In the example, it is inferred that the concept \texttt{special\_occasion} causes shiok as it shares the same set of semantic features with \texttt{hari\_raya} and \texttt{birthday}}{tbl:cumulative_analogy}
\end{table}

To perform inference on multiple matrices, blending is the most widely used technique. It allows multiple matrices to be combined in a single matrix, basing on the overlap between these matrices. 
The new matrix is rich in information and contains much of the information shared by the two original matrices. By means of the singular value decomposition on the new matrix, new connections are formed in source matrices based on the shared information and overlap between them. This method enables creation of a new resource, which is a combination of multiple resources representing different kinds of knowledge.
In order to build a suitable knowledge base for affective reasoning, we applied the blending technique to ConceptNet, AffectNet and Singlish AffectNet. 
First, we represented AffectNet as a directed graph, similarly to ConceptNet. For example, the concept birthday party has the associated emotion joy; we considered birthday party and joy as two nodes, and added an assertion HasProperty on the edge directed from the node birthday party to the node joy. 
Next, we converted the three graphs, ConceptNet and AffectNet and Singlish AffectNet, to sparse matrices to blend them. After blending the two matrices, we performed TSVD on the resulting matrix, to discard those components that represent relatively small variations in the data. 
We kept only 100 components of the blended matrix to obtain a good approximation of the original matrix. 

\subsubsection{Obtaining Polarity Values of Singlish Concepts}
The proposed framework is designed to receive as input a natural language concept represented according to an $M$-dimensional space and to predict the corresponding sentic levels for the four affective dimensions involved: Pleasantness, Attention, Sensitivity, and Aptitude. 
The dimensionality $M$ of the input space stems from the specific design of Singlish AffectiveSpace. As for the outputs, in principle each affective dimension can be characterized by an analog value in the range $[-1,1]$, which represents the intensity of the expressed or received emotion. 

Indeed, those analog values are eventually remapped to obtain six different sentic levels for each affective dimension.
The categorization framework spans each affective dimension separately, under the reasonable assumption that the various dimensions map perceptual phenomena that are mutually independent. As a result, each affective dimension is handled by a dedicated ELM \cite{huanew}, which addresses a regression problem. 

\begin{figure}[t]
	\centering
	\caption{\small{Blending of Singlish AffectNet with ConceptNet and English AffectNet }}
	\includegraphics[width=\linewidth]{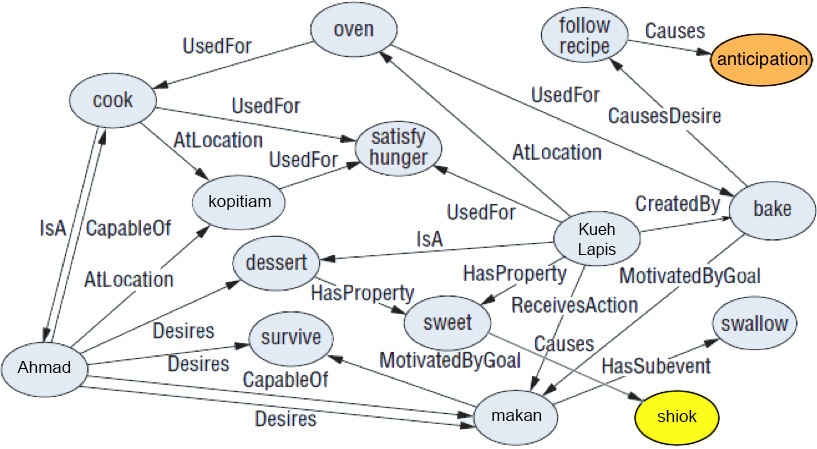}
	\label{figure:singaffect}
\end{figure}

\begin{figure}[t]
	\centering
	
	\caption{\small{The ELM-based framework for describing common-sense concepts in terms of the four Hourglass model’s dimensions }}
	\vspace{3mm} 
	\includegraphics[width=7cm]{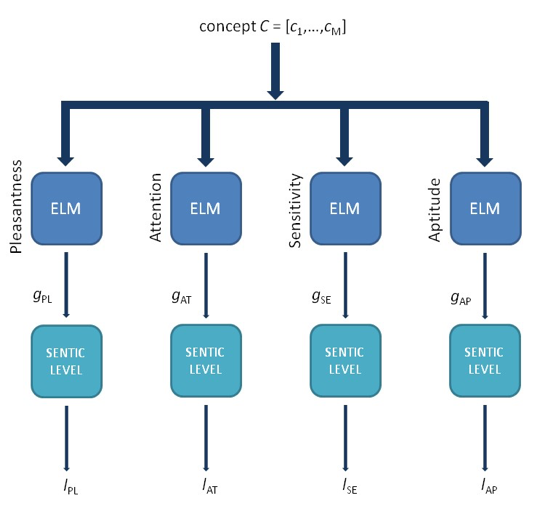}
	\label{figure:ELM}
\end{figure}

Thus, each ELM-based predictor is fed by the $M$-dimensional vector describing the concept and yields as output the analog value that would eventually lead to the corresponding sentic level. 
Figure \ref{figure:ELM} provides the overall scheme of the framework; here, $g_X$ is the level of activation predicted by the ELM and $l_X$ is the corresponding sentic level.
In theory, one might also implement the framework showed in Figure \ref{figure:ELM} by using four independent predictors based on a multi-class classification schema. In such a case, each predictor would directly yield as output a sentic level out of the six available. 
However, two important aspects should be taken into consideration. First, the design of a reliable multi-class predictor is not straightforward, especially when considering that several alternative schemata have been proposed in the literature without a clearly established solution. Second, the emotion categorization scheme based on sentic levels stem from an inherently analog model, i.e., the Hourglass of Emotions. This ultimately motivates the choice of designing the four prediction systems as regression problems.
In fact, the framework schematized in Figure \ref{figure:ELM} represents an intermediate step in the development of the final emotion categorization system.
One should take into account that every affective dimension can in practice take on seven different values: the six available sentic levels plus a `neutral' value, which in theory correspond to the value $G(0)$ in the Hourglass model. 

In practice, though, the neutral level is assigned to those concepts that are characterized by a level activation that lies in an interval around $G(0)$ in that affective dimension.
Therefore, the final framework should properly manage the eventual seven-level scale. 

\section{Conclusion}
In this work, we proposed the construction of a concept-level Singlish sentiment lexicon using the English SenticNet framework. In future, we plan to extract 1m Singlish concepts from various Singlish blogs and websites and employ the same framework. The proposed framework could also be improved by using different supervised classifiers.

\bibliographystyle{splncs03}
\bibliography{ref-1}
\end{document}